\documentclass[10pt,twocolumn,letterpaper]{article}

\usepackage{cvpr}
\usepackage{times}
\usepackage{epsfig}
\usepackage{graphicx}
\usepackage{amsmath}
\usepackage{amssymb}

\usepackage[breaklinks=true,bookmarks=false]{hyperref}
\usepackage{authblk}
\cvprfinalcopy 

\def\cvprPaperID{****} 

\begin{document}


\title{Context Aware Road-user Importance Estimation (iCARE)}

\author[1,2]{Alireza Rahimpour\thanks{Work performed while interning at Honda Research Institute, USA.}}
\author[2]{Sujitha Martin}
\author[2]{Ashish Tawari}
\author[1]{Hairong Qi}

\affil[1]{\normalsize{
Department of Electrical Engineering and Computer Science}\\

\normalsize{
University of Tennessee, Knoxville, TN. $ \{ $arahimpo, hqi$ \} $@utk.edu}}
\affil[2]{\normalsize{
Honda Research Institute, Mountain View, CA.$ \{ $smartin, atawari$ \} $@honda-ri.com} }

\date{}                     
\renewcommand\Affilfont{\itshape\small}


\maketitle

\begin{abstract}
  Road-users are a critical part of decision-making for both self-driving cars and driver assistance systems. Some road-users, however, are more important for decision-making than others because of their respective intentions, ego-vehicle's intention and their effects on each other. In this paper, we propose a novel architecture for road-user importance estimation which takes advantage of the local and global context of the scene. For local context, the model exploits the appearance of the road users (which captures orientation, intention, etc.) and their location relative to ego-vehicle. The global context in our model is defined based on the feature map of the convolutional layer of the module which predicts the future path of the ego-vehicle and contains rich global information of the scene (e.g., infrastructure, road lanes, etc.), as well as the ego-vehicle's intention information. Moreover, this paper introduces a new data set of real-world driving, concentrated around intersections and includes annotations of important road users. Systematic evaluations of our proposed method against several baselines show promising results.

\end{abstract}

\section{Introduction}

In real-world driving, at any given time, there can be many road-users in the ego-vehicle's vicinity. Some road-users directly affect ego-vehicle's behavior (i.e. brake, steer), while some could be a potential risk and others who do not pose a risk at this time or in the near future (as illustrated in Figure \ref{fig:fig1}). The ability to discern how important or relevant any given road user is to an ego-vehicle's decision is vital for building trust with human drivers or passengers, transparency with law makers, promoting human-centric thought process, etc., for both driver assistance systems and self-driving cars. In this paper, we propose to estimate road-user importance based on visually guided information; here an important road-user is defined as that which is or will affect the ego-vehicle's dynamics (e.g. brake, steer, accelerate).

Given a single image of a driving scene, visually, humans have an unparalleled ability to determine which road users are affecting or likely to affect the ego-vehicle's behavior. Humans leverage information such as traffic rules, intended path of ego-vehicle, potential trajectory of road participants, location, etc. As many of these information can be inferred from the image, we propose a method to estimate road-user importance by taking advantage of local and global context. We call this method, Context Aware Road-user Importance Estimation (iCARE). iCARE takes an image from $3$ cameras in front of the ego-vehicle as input and outputs the importance of road users as well as the future path of the ego-vehicle. 

In iCARE, local context is represented by appearance of road users (which captures orientation, intention, etc.) and their location relative to ego-vehicle. Global context is represented with the feature map of the last convolutional layer (denoted as global context in Figure \ref{fig:fig}) of the model which is trained to predict the future path of the ego-vehicle. The global context has rich global information of the scene such as infrastructure, road lanes, etc. Moreover, to some extent this can be considered intention-based context because this same representation is used to predict future path of ego-vehicle. To this end, the main contributions of this paper are as following:

\begin{figure*}[th]
	\centering
	\includegraphics[width=1\linewidth, height=0.2\textheight]{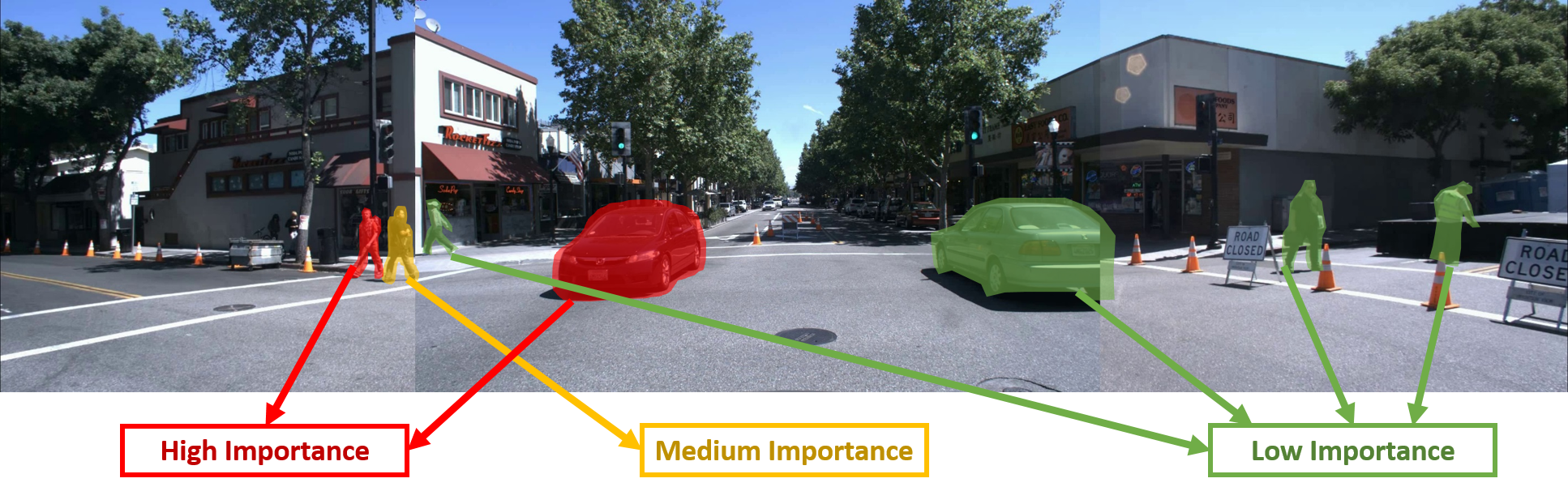}
	\caption{Illustration of an ideal road-user importance estimation during a left turn maneuver. In a driving scenario, there can be many road-users. However, when given an ego-vehicle's path, some road-users are more important for decision-making.} 
	\label{fig:fig1}
\end{figure*}

\begin{itemize}
 \item Designing a new image-based framework for estimating the importance level of road users.  

\item Proposing a novel context aware architecture and a new way of representing the global context of the scene based on predicting the intention (future path) of the ego-vehicle. 

\item Collecting a new data set of on-road driving concentrated around intersections with human annotated road-user importance.

\item Systematic quantitative and qualitative evaluation of the proposed method against several baselines.
\end{itemize}

\subsection{Related Works}

Recently many efforts have been devoted to development of vehicles with higher level of autonomy based on scene understanding and saliency map estimation.
For instance, driver's gaze has been widely studied for determining saliency map and intention prediction relying only on fixation maps \cite{pugeault2015much}. \cite{underwood2011decisions} inspects the driver's attention specifically towards pedestrians and motorbikes, and exploits object saliency. 
In \cite{palazzi2017predicting}, a computer vision based model is proposed to predict saliency by conducting a data-driven study on drivers' gaze fixations.
However, driver's gaze is not always a valid indication of saliency since the driver might look at many unimportant objects in the scene as well. 

Different from our proposed method, prediction of important objects is also studied by 
\cite{kuen2016recurrent, li2016deepsaliency}. \cite{kuen2016recurrent} uses recurrent attention and convolutional-deconvolutional network to tackle the salient object detection problem.
Furthermore, the proposed model in \cite{li2016deepsaliency} takes a strategy for encoding the underlying saliency prior information, and then sets up a multi-task learning scheme for exploring the intrinsic correlations between salient object detection and semantic image segmentation. However, these methods are not applicable to road user importance estimation in driving scenario which highly depends on the ego-vehicle's intention and its interaction with other road users.

\begin{figure*}[ht!]
	\centering
	\includegraphics[width=1\linewidth]{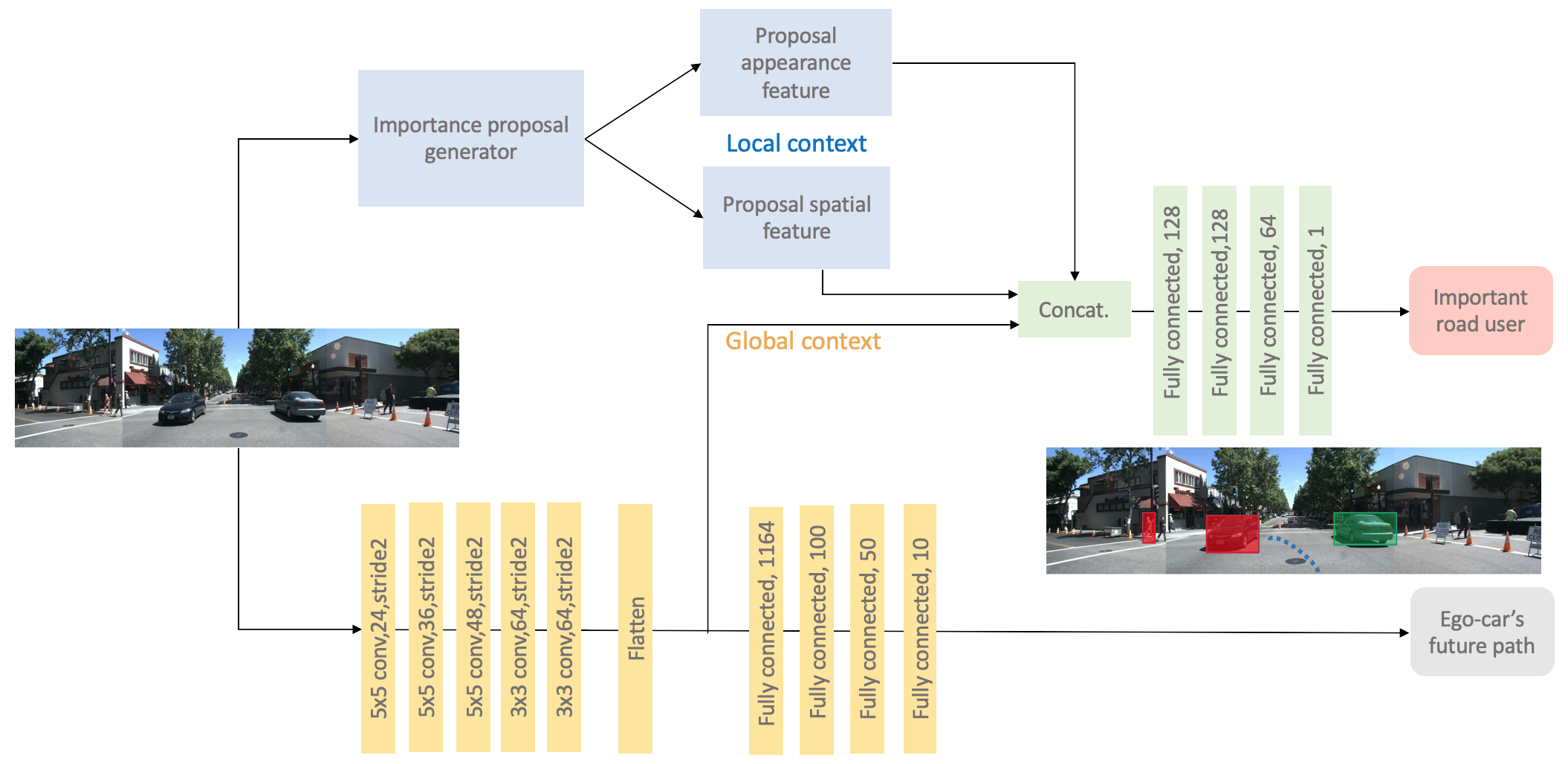}
	\caption{The iCARE model exploits local (i.e. appearance, location) of road users and global (i.e. intention based context) of the scene to estimate importance of respective road users.}
	\label{fig:fig}
\end{figure*}

Another approach for solving the saliency estimation problem in autonomous driving is using sensor-based methods.
LiDAR \cite{halterman2010velodyne}, radars, lasers and sonars \cite{park2003novel} are popular sensors to detect surrounding objects in autonomous systems. For instance, \cite{sheu2007ddas} uses smart antennas and proposes a distance awareness system for important object estimation. 
The model proposed by \cite{chen2017multi} combines the front view of the LiDAR point cloud with region-based features from the bird's eye view for 3D object detection. 
However, the salient objects are not necessarily the nearest object (e.g. nearest object like a parked car may not pose as much a threat as a pedestrian intending to cross ego-vehicle's path further down the road).
Therefore, contextual information is essential for practical autonomous driving systems. 
For more details about the history of using different sensors and methods for autonomous driving systems please refer to \cite{janai2017computer}.
 
Different from recent works based on estimating a general saliency map of the scene (i.e., a heat map which gives each pixel a relative value of its level of saliency) \cite{TawariITSC2018, rahimpour2017person}, our proposed method is able to specifically estimate the importance level of the road users based on the scene context and ego-vehicle's intention. Furthermore, unlike the works based on estimating the driver's gaze fixation map \cite{CorniaITMP2018}, in this paper we propose a road user importance estimation method based on human-centric importance annotation.

\section{Method}


An overview of the proposed Context Aware Road user importance Estimation (iCARE) model is shown in Figure \ref{fig:fig}. There are two main stages in the proposed model. First, an important road user proposal generator provides potentially important road user proposals and then in the second stage, context is incorporated into the system in order to estimate the importance level of road users. The next subsections describe these stages in more details.

\subsection{Important Road user Proposal Generation}

In this stage, Faster R-CNN \cite{ren2015faster} is exploited in order to generate the potentially important road users by training as a 1-class classification on important road-users. The important road user proposal generation is performed by using the Region Proposal Network (RPN) which predicts object proposals and at each spatial location, the model predicts a class-agnostic objectness score and a bounding box refinement for anchor boxes of multiple scales and aspect ratios. Using non-maximum suppression with an intersection-over-union (IoU) threshold, the top box proposals are selected. Then, region of interest (RoI) pooling is used to extract a fixed-size feature map for each box proposal. These feature maps then go through the final fully connected layers where the class label (i.e, important road user) and bounding box refinements for each box proposal are obtained. This stage of our model is designed to select more likely candidates of importance, where further consideration into context is necessary to accept or reject the proposal.

\subsection{Context Aware Representation and Fusion}

After selecting potentially important road users in the first stage, in the next stage, iCARE fuses the local (i.e. appearance, location) and global context towards estimating importance of road users. 

\textbf{Local Appearance Feature}

The appearance feature of the road users contains very rich information about orientation, dynamics, intention, distance, etc. of the road users.  
In this work, we use the Inception-ResNet-V2 \cite{szegedy2017inception} model as feature extractor in conjunction with the road user proposal generator model. To train the important road user proposal generator model, we first initialize it with Faster R-CNN trained on COCO object detection dataset \cite{lin2014microsoft} and then train it based on the important road user annotations in our data set. 
To generate the appearance feature for each potential important road user proposal, we take the final output of the model and select all bounding boxes where the probability of belonging to the ``important" class exceeds a confidence threshold. For each selected proposal, the appearance feature is defined as the output of the region of interest (RoI) pooling layer for that bounding box. 

\textbf{Location Feature}

Road users with different sizes and distances to the ego-vehicle have different attributes which can make them better distinguishable. In our model, for each proposal of important road user, we consider a $4D$ vector as the location feature $f_{loc}$ which is defined as:
\begin{equation}
f_{loc} = [((x_{max} + x_{min})/2, y_{max}), h, w],
\end{equation}
where $((x_{max} + x_{min})/2, y_{max})$ is the coordinate of the middle bottom point of each proposal bounding box and $h$ and $w$ are the height and width of the bounding box, respectively. The location feature helps the system to learn the correlation between proximity, mass and importance. 

\textbf{Intention-based (Global) Context}

Intention of the ego-vehicle plays a major role in estimating the importance of road users. For instance, if the ego-vehicle's intention is to make a left turn in an intersection, then road users on the right side of the intersection may be considered relatively less important road users. In order to incorporate the intention of the ego-vehicle, we design a model (inspired by \cite{bojarski2016end}, where the model learns a mapping between an image and instantaneous steering angle) which takes as input a single image of the scene and predicts a $10D$ vector of the future path of the car. 
The future path vector is constructed of $10$ steering angle values, representing the next $10$ spatial steps of the car with $1$-meter equal spacing. 
The model for predicting the future path (shown in blue in Figure \ref{fig:fig}) consists of convolution layers followed by fully connected layers with batch normalization and drop out layers in between. The flatten feature of the last convolution layer is used as the context feature.   

\textbf{Feature Fusion}

In this part of the model, the local (i.e., appearance and location) and global (i.e., intention-based context) features are concatenated together and followed by $4$ fully connected layers to estimate importance of a road user (shown in green in Figure \ref{fig:fig}). To study the effect of intention and context, we consider a version of our model were instead of context, the $10D$ ground truth future path vector is used as an input to the model and the results are compared. 
Furthermore, an ablation study is performed on combinations of the features (i.e. appearance, spatial, intention context and the future path of the car as input) which will be elaborated on in the next section.

\section{Experiments}
\subsection{Data set}

The data set used for training and evaluation of the proposed iCARE model is collected around urban intersections and includes the aligned-view images that are generated by putting together (and aligning) the images from three cameras (i.e., left, center and right views). The data set consists of $6$ hours of driving including around $2.7$ hours of intersections. The data is collected from driving on the streets of Mountain View and Sunnyvale in California. There are $743$ total intersection segments which include $307908$ total frames. 

\begin{figure}[h]
	\centering
	\includegraphics[width=1\linewidth , height=0.3\textheight]{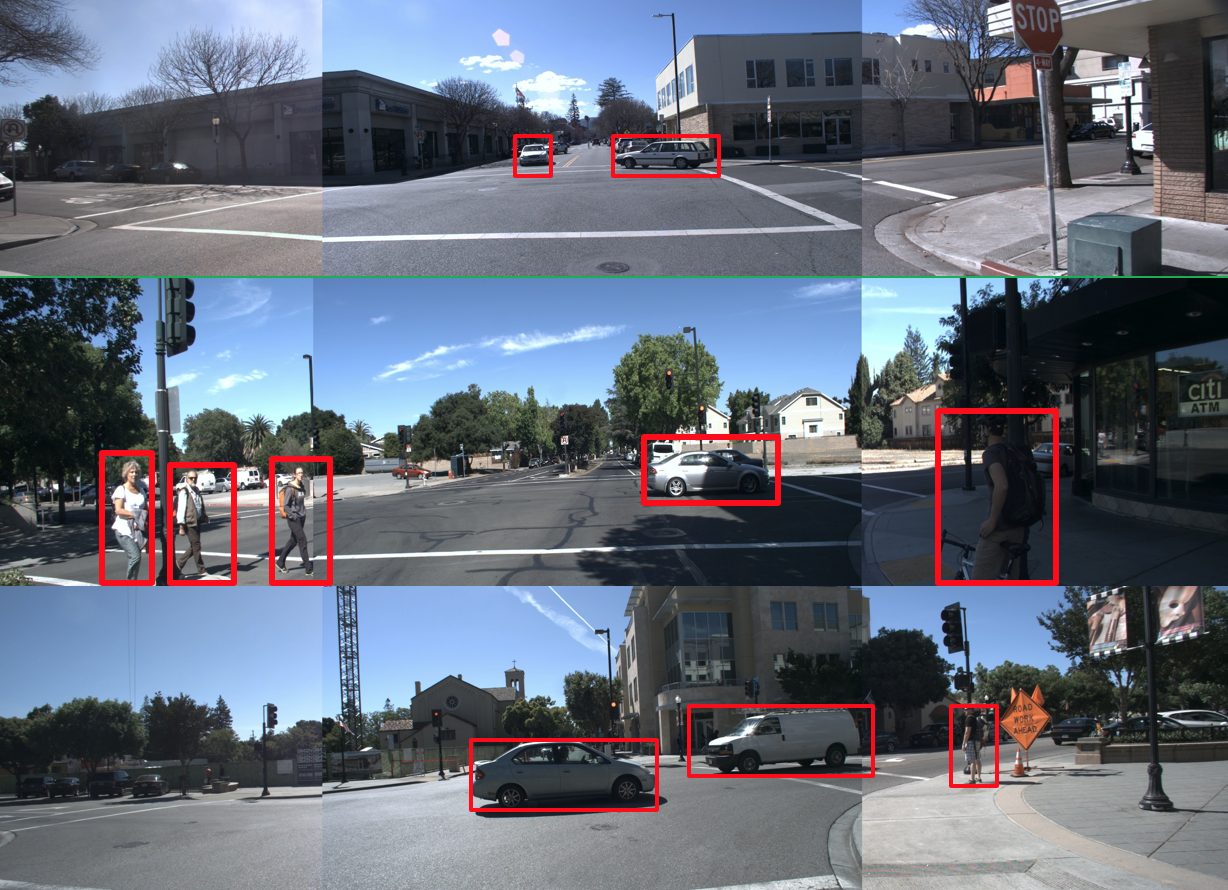}
	\caption{Examples of the images and annotations in our data set}
	\label{fig:fig3}
\end{figure}

The important road users in image sequences are annotated every $30$ frames (i.e., $1$ annotation per second). The annotations are generated by human annotators who are given the video sequences and instructions about the driving rules.  
There are $9995$ total frames of annotations, of which $6924$ total road users are annotated as important. An annotated image may include between zero to five important road users.
Some examples from the data set are illustrated in Figure \ref{fig:fig3}. 

The data is split to training set and testing set with $13624$ and $4749$ images, respectively. Only images with annotation (i.e., with at least one important road user in them) are used for training, but testing is performed on all images in the test set.

\subsection{Implementation Details}

In our model, the aligned view images are re-sized from their original size (i.e., $4394 \times 1100$) to $1024 \times 275$ before going into the deep neural network model. Tensorflow v1.4 is used as our deep learning framework on a Tesla V100-SXM2 NVIDIA GPU with $32$ gigabytes of memory. 
 
For the intention-based context extractor branch, we use $5$ convolution layers followed by $4$ fully connected layers. Batch normalization \cite{ioffe2015batch} is used for faster training and also there are drop out layers (with keep-prob $= 0.6$) between the fully connected layers to avoid over fitting. The intention-based context branch is trained and tested based on the same data split used for important road user proposal generation. The last convolution layer of this model is extracted and flattened to $1164D$ vector and used as the context feature for each image. Mean Square Error (MSE) is used as loss function for training of this part of the model.  

iCARE concatenates the local features from the proposal generator and the context features and then there are $4$ fully connected layers with $128$, $128$, $64$ and $1$ neurons with batch normalization and drop out between the layers. Binary cross entropy is used as the loss function for the final classification step (i.e., important vs not important road users). Adam optimizer \cite{kingma2014adam} with $\beta_1 = 0.9 $ and $\beta_2 = 0.99 $ and learning rate of $0.01 $ is used for optimization of the loss functions in all parts of our model. Moreover, the Relu \cite{nair2010rectified} is used as non-linearity throughout our model.

\subsection{Evaluation and Results}

Since the number of not important road users is larger than number of important road users in our data set, we need to deal with data imbalance problem in training our model. In fact, in the training set there are $4699$ important samples and $8925$ not important samples.                                                                                                     
In order to solve the data imbalance problem we assign appropriate weights for the loss terms of each class (i.e., $1:2$).
Furthermore, data imbalance causes the classification accuracy metric (i.e., unweighted accuracy) to not be able to precisely estimate the performance of the model. Hence, we use the precision-recall curve and $F1$ score to evaluate our model. The $F1$ score is the harmonic average of the precision and recall, and it reaches its best value at $1$ (perfect precision and recall) and worst at $0$.  
 
\begin{figure}[ht!]
	\centering
	\includegraphics[width=1\linewidth , height=0.33\textheight]{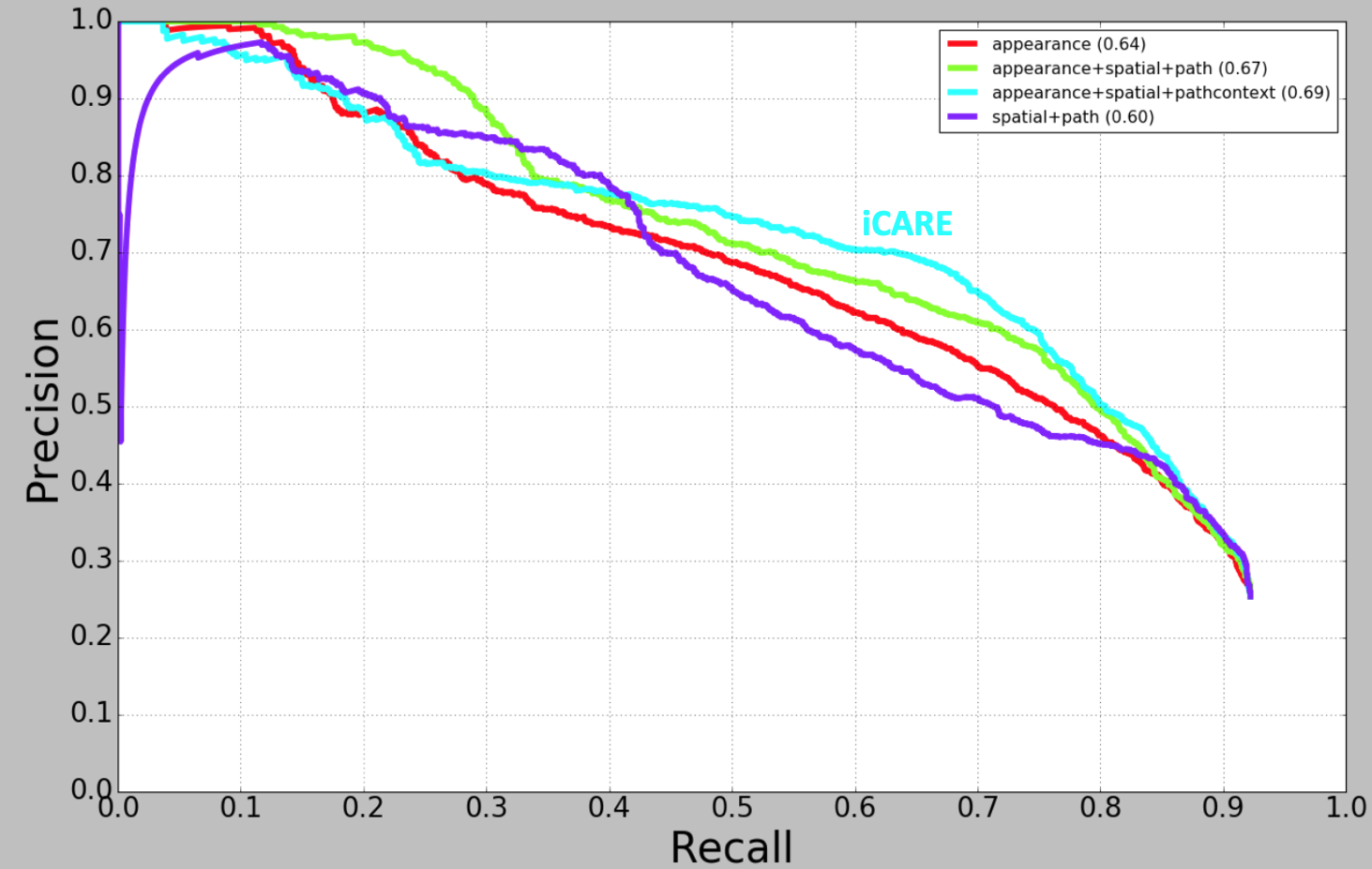}
	\caption{Precision-recall curves (and $F1$ scores) for different experiment settings. Best viewed in color.}
	\label{fig:fig4}
\end{figure}
The precision-recall curves for different experiment settings are shown in Figure \ref{fig:fig4}. The $F1$ score is shown in parenthesis for each experiment, as well. In Figure \ref*{fig:fig4}, the red curve (denoted as appearance ($0.65$)) is the setting where only the appearance feature taken from the important road user proposal is used to estimate the road user importance. 
The green curve (denoted as: appearance + spatial + path ($0.67$) in Figure \ref*{fig:fig4}) corresponds to the setting where the future path is used as an input to the model and is concatenated with appearance and location features. This combination of the features yields an $F1$ score of $0.67$ which is $3\%$ higher than using only the appearance feature.  
The light blue curve which achieves the best performance ($F1 = 0.69$) is the result of the full iCARE model which exploits the intention-based context of the scene along with the local features (i.e., appearance and location features). This result shows the importance of exploiting intention-based global context representation of the scene. 

Moreover, the magenta curve illustrates the experiment setting where the appearance of the road users is not exploited but only location and future path (as input) are used. The results show that this setting achieves the lowest performance of $F1 = 0.60$ as it is expected. It shows the importance of using the appearance features which has rich information about the orientation, type, etc. of the road users. It is worth noting that the reported results in this section are all based on testing on all the images in the testing set where even images without any ground truth annotation are also considered for testing. Testing only on images with annotation achieves results with around $10\%$ improvement compared to the reported results. 

\begin{figure}[h]
	\centering
	\includegraphics[width=1\linewidth , height=0.35\textheight]{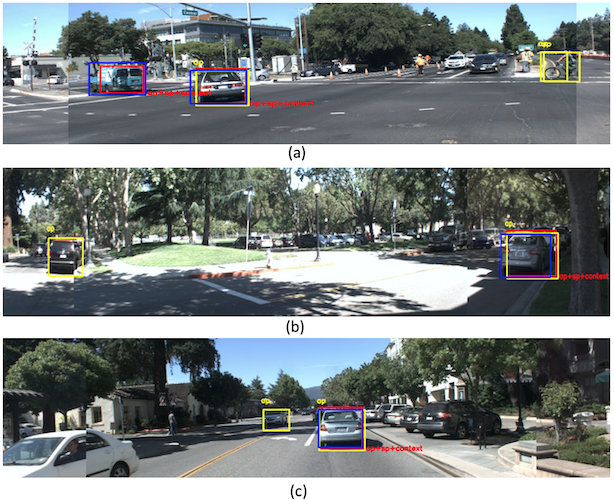}
	\caption{Examples of performance comparison of using appearance feature only (yellow) vs iCARE (red) and ground truth (blue).}
	\label{fig:fig5}
\end{figure}

The qualitative results of road user importance estimation using our proposed model are illustrated in Figure \ref{fig:fig5}. In this Figure, the blue bounding boxes correspond to the ground truth annotation and the yellow bounding boxes correspond to the estimation of the model when using only the appearance feature. The red bounding boxes show the result of our iCARE model which considers the intention based context of the scene as well as the appearance and location features of the road users. It can be seen in Figure \ref{fig:fig5}-(a), when using only the appearance feature, model gives false positive estimations for the cyclist on the right hand side of the intersection. Moreover, for the car on the most left hand side, only the iCARE can successfully detect the important road users based on the learned intention from the scene context. Furthermore, Figure \ref{fig:fig5}-(b) and (c) show two very common cases in our test set that incorporating the ego-vehicle's intention based context helps to get rid of the false positive estimations when using only the appearance of the road users. 
 
\begin{figure}[h]
	\centering
	\includegraphics[width=1\linewidth, height=0.30\textheight]{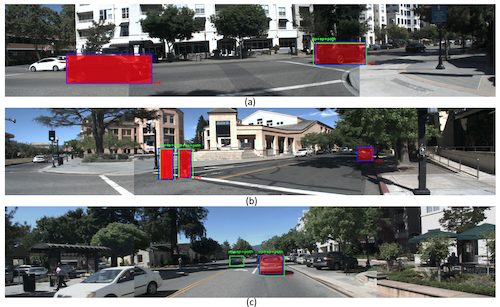}
	\caption{Comparison of performance of iCARE model (red) vs fusion of appearance, spatial and input future path features (green). The blue bounding boxes show the ground truth annotations for important road users. The intensity of the red color shows the level of importance of each road user estimated by iCARE. Best viewed in color.}
	\label{fig:fig6}
\end{figure}

\begin{figure}
    \centering
    \includegraphics[width=1\linewidth, height=0.25\textheight]{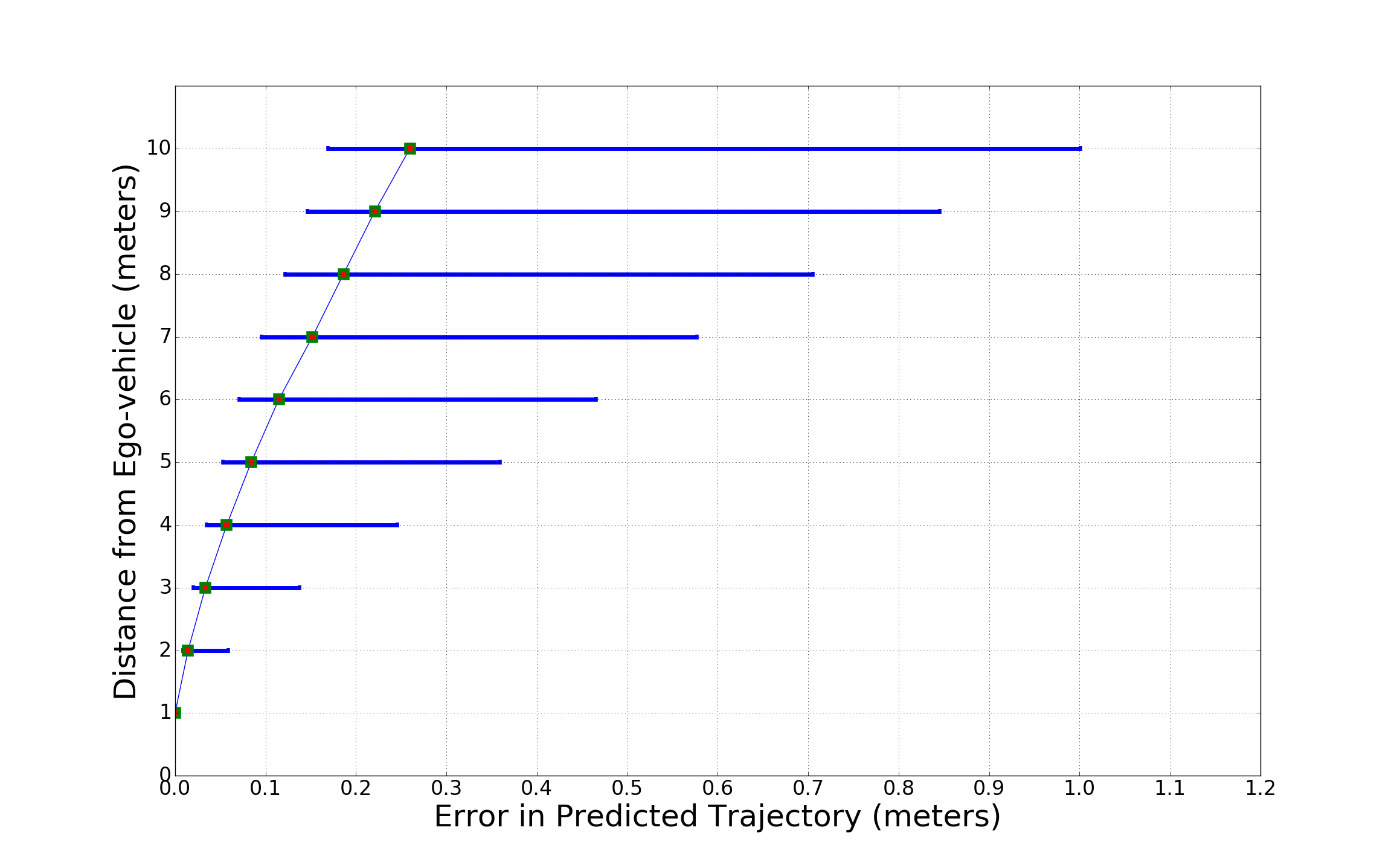}
    \caption{Ego-vehicle future path prediction error versus distance from ego-vehicle.}
    \label{fig:my_label}
\end{figure}

In one experiment we also compare the performance of our model when using the future path of the ego-vehicle as input to the model versus when the model uses the context from path prediction (i.e., iCARE). Figure \ref{fig:fig6} shows some examples of this experiment. In this Figure the red layover color demonstrates the output of the iCARE model and the intensity of the red color illustrates how important that road user is based on the prediction of our model. The blue bounding box corresponds to the ground truth and the green bounding box corresponds to the experiment setting where the future path of the ego-vehicle has been used as input feature (along with the appearance and location feature). It can be observed that using the $10D$ future path vector as input is not as effective as using the future path context. For instance, it can be observed in Figure \ref{fig:fig6}-(a)-(b) that when using the future path as input, model can not estimate the important road users properly, and also it sometimes leads to false positives as it is shown in Figure \ref{fig:fig6}-(c).
Moreover, the estimation error in predicting the future path of the car is shown in Figure \ref{fig:my_label}. It can be observed that the estimation error increases as the distance to the ego vehicle increases. 

\begin{figure}[h]
	\centering
	\includegraphics[width=1\linewidth, height=0.3\textheight]{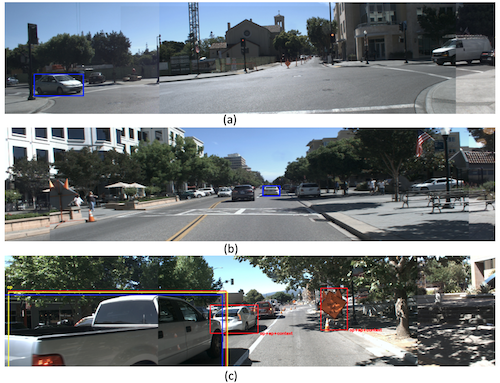}
	\caption{Three examples of failure cases of iCARE model estimations. (red: iCARE estimation, blue: ground truth, yellow: using only appearance feature for importance estimation.)}
	\label{fig:fig7}
\end{figure}

Some examples of our model's failure estimations are shown in Figure \ref{fig:fig7}. For instance in Figure \ref{fig:fig7}-(a), the iCARE model is not able to detect the left car in the intersection. This is mainly due to lack of road user's intention information (i.e. if the white car's intention is to turn right then it is not indeed important, but if it is going straight it should be considered as important). Moreover, when some road users are very far away from the ego-vehicle (Figure \ref{fig:fig7}-(b)), iCARE estimates them as not important. Another failure case is due to mis-detection and other unavoidable causes (Figure \ref{fig:fig7}-(c)). Interestingly, sometimes estimations of the iCARE does make sense even though those road users have not been annotated as important (e.g., traffic sign in Figure \ref{fig:fig7}-(c)).

\begin{figure}[h!]
	\centering
	\includegraphics[width=1\linewidth]{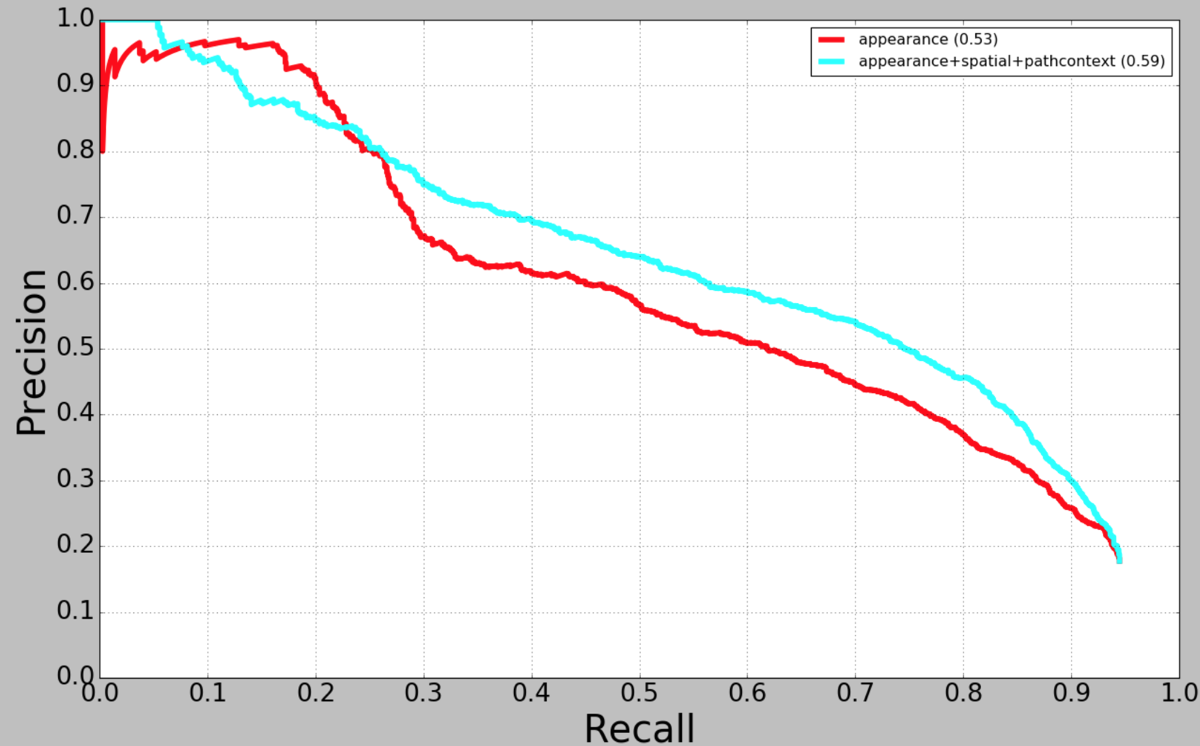}
	\caption{Precision-recall curves for iCARE (light blue) and the baseline (red) when trained on the first annotation and test on the second annotation.}
	\label{fig:fig8}
\end{figure}

In another experiment, we investigate the subjectivity issue in road user importance estimation. In fact, even though most drivers agree on the obvious important road users (e.g., a pedestrian in front of the moving ego-vehicle, etc.), different drivers might have different opinion about importance of some of the road users. In order to study this subjectivity, annotation from a different annotator is used to test our model. 
The performance of iCARE versus appearance-based baseline when trained with first annotation and tested with second annotation is shown in Figure \ref{fig:fig8}. This Figure shows precision-recall curves of the proposed iCARE model (shown in light blue) with $F1 =0.59$ and appearance-based baseline (shown in red) with $F1 = 0.53$. It can be observed that even though the iCARE model achieves lower accuracy (compared to train and test with same annotation), it still works fairly well and has a consistent behavior with our previous results.  

\section{Conclusion Remarks}

In this paper, we investigated the effect of ego-vehicle's intention and its context on estimating road users importance using only images taken from $3$ cameras in front of the car. The proposed iCARE model estimates the important road users based on a $2$-stage recognition framework, where the first stage generates important road user proposals using an importance-guided training scheme. In the second stage, model estimates the importance level of the road user proposals by taking into account the location and intention context information. Our future work is to incorporate the intention of the road users into our model which plays an important role in determining which road user is important. Furthermore, incorporating other contextual information (e.g., depth, motion, etc.) can be an interesting line of future research for road user importance estimation. 

{
\small
\bibliographystyle{ieee}
\bibliography{egbib}

\begin{thebibliography}{10}\itemsep=-1pt

\bibitem{bojarski2016end}
M.~Bojarski, D.~Del~Testa, D.~Dworakowski, B.~Firner, B.~Flepp, P.~Goyal, L.~D.
  Jackel, M.~Monfort, U.~Muller, J.~Zhang, et~al.
\newblock End to end learning for self-driving cars.
\newblock {\em arXiv preprint arXiv:1604.07316}, 2016.

\bibitem{chen2017multi}
X.~Chen, H.~Ma, J.~Wan, B.~Li, and T.~Xia.
\newblock Multi-view 3d object detection network for autonomous driving.
\newblock In {\em IEEE CVPR}, volume~1, page~3, 2017.

\bibitem{CorniaITMP2018}
M.~Cornia, L.~Baraldi, G.~Serra, and R.~Cucchiara.
\newblock Predicting human eye fixations via an lstm-based saliency attentive
  model.
\newblock {\em IEEE Transactions on Image Processing}, 27(10):5142--5154, Oct
  2018.

\bibitem{halterman2010velodyne}
R.~Halterman and M.~Bruch.
\newblock Velodyne hdl-64e lidar for unmanned surface vehicle obstacle
  detection.
\newblock In {\em Unmanned Systems Technology XII}, volume 7692, page 76920D.
  International Society for Optics and Photonics, 2010.

\bibitem{ioffe2015batch}
S.~Ioffe and C.~Szegedy.
\newblock Batch normalization: Accelerating deep network training by reducing
  internal covariate shift.
\newblock {\em arXiv preprint arXiv:1502.03167}, 2015.

\bibitem{janai2017computer}
J.~Janai, F.~G{\"u}ney, A.~Behl, and A.~Geiger.
\newblock Computer vision for autonomous vehicles: Problems, datasets and
  state-of-the-art.
\newblock {\em arXiv preprint arXiv:1704.05519}, 2017.

\bibitem{kingma2014adam}
D.~P. Kingma and J.~Ba.
\newblock Adam: A method for stochastic optimization.
\newblock {\em arXiv preprint arXiv:1412.6980}, 2014.

\bibitem{kuen2016recurrent}
J.~Kuen, Z.~Wang, and G.~Wang.
\newblock Recurrent attentional networks for saliency detection.
\newblock In {\em Proceedings of the IEEE Conference on Computer Vision and
  Pattern Recognition}, pages 3668--3677, 2016.

\bibitem{li2016deepsaliency}
X.~Li, L.~Zhao, L.~Wei, M.-H. Yang, F.~Wu, Y.~Zhuang, H.~Ling, and J.~Wang.
\newblock Deepsaliency: Multi-task deep neural network model for salient object
  detection.
\newblock {\em IEEE Transactions on Image Processing}, 25(8):3919--3930, 2016.

\bibitem{lin2014microsoft}
T.-Y. Lin, M.~Maire, S.~Belongie, J.~Hays, P.~Perona, D.~Ramanan,
  P.~Doll{\'a}r, and C.~L. Zitnick.
\newblock Microsoft coco: Common objects in context.
\newblock In {\em European conference on computer vision}, pages 740--755.
  Springer, 2014.

\bibitem{nair2010rectified}
V.~Nair and G.~E. Hinton.
\newblock Rectified linear units improve restricted boltzmann machines.
\newblock In {\em Proceedings of the 27th international conference on machine
  learning (ICML-10)}, pages 807--814, 2010.

\bibitem{palazzi2017predicting}
A.~Palazzi, D.~Abati, S.~Calderara, F.~Solera, and R.~Cucchiara.
\newblock Predicting the driver's focus of attention: the dr (eye) ve project.
\newblock {\em arXiv preprint arXiv:1705.03854}, 2017.

\bibitem{park2003novel}
S.~J. Park, T.~Y. Kim, S.~M. Kang, and K.~H. Koo.
\newblock A novel signal processing technique for vehicle detection radar.
\newblock In {\em Microwave Symposium Digest, 2003 IEEE MTT-S International},
  volume~1, pages 607--610. IEEE, 2003.

\bibitem{pugeault2015much}
N.~Pugeault and R.~Bowden.
\newblock How much of driving is preattentive?
\newblock {\em IEEE Transactions on Vehicular Technology}, 64(12):5424--5438,
  2015.

\bibitem{rahimpour2017person}
A.~Rahimpour, L.~Liu, A.~Taalimi, Y.~Song, and H.~Qi.
\newblock Person re-identification using visual attention.
\newblock In {\em 2017 IEEE International Conference on Image Processing
  (ICIP)}, pages 4242--4246. IEEE, 2017.

\bibitem{ren2015faster}
S.~Ren, K.~He, R.~Girshick, and J.~Sun.
\newblock Faster r-cnn: Towards real-time object detection with region proposal
  networks.
\newblock In {\em Advances in neural information processing systems}, pages
  91--99, 2015.

\bibitem{sheu2007ddas}
S.-T. Sheu, J.-S. Wu, C.-H. Huang, Y.-C. Cheng, L.~Chen, et~al.
\newblock Ddas: Distance and direction awareness system for intelligent
  vehicles.
\newblock {\em Journal of information science and engineering}, 23(3):709--722,
  2007.

\bibitem{szegedy2017inception}
C.~Szegedy, S.~Ioffe, V.~Vanhoucke, and A.~A. Alemi.
\newblock Inception-v4, inception-resnet and the impact of residual connections
  on learning.
\newblock In {\em AAAI}, volume~4, page~12, 2017.

\bibitem{TawariITSC2018}
A.~Tawari, P.~Mallela, and S.~Martin.
\newblock Learning to attend to salient targets in driving videos using fully
  convolutional rnn.
\newblock In {\em 2018 21st International Conference on Intelligent
  Transportation Systems (ITSC)}, pages 3225--3232, Nov 2018.

\bibitem{underwood2011decisions}
G.~Underwood, K.~Humphrey, and E.~Van~Loon.
\newblock Decisions about objects in real-world scenes are influenced by visual
  saliency before and during their inspection.
\newblock {\em Vision research}, 51(18):2031--2038, 2011.

\end{thebibliography}
}

\end{document}